\documentclass[a4paper,fleqn]{cas-sc}

\usepackage[authoryear,longnamesfirst]{natbib}
\usepackage{amssymb}
\usepackage{amsmath}
\usepackage{booktabs}
\usepackage{multirow}
\usepackage{subcaption}

\usepackage{comment}

\def\tsc#1{\csdef{#1}{\textsc{\lowercase{#1}}\xspace}}
\tsc{WGM}
\tsc{QE}
\tsc{EP}
\tsc{PMS}
\tsc{BEC}
\tsc{DE}

\begin{document}
\let\WriteBookmarks\relax
\def\floatpagepagefraction{1}
\def\textpagefraction{.001}
\shorttitle{A Statistical Multi-Objective Framework for Assessing Sensitivity of  Radiomic AI Models to Acquisition Parameters}
\shortauthors{D.Gil, I.Sanchez, C.Sanchez}

\title [mode = title]{A Statistical Multi-Objective Framework for Assessing Sensitivity of  Radiomic AI Models to Acquisition Parameters}

\tnotetext[1]{This document is the results of the research
   project funded by funded by MICIU/AEI/ 10.13039/501100011033 and by ERDF/EU, and Agència de Gestió d'Ajuts Universitaris i de Recerca grant numbers 2025 PROD 00239 and CERCA Programme / Generalitat de Catalunya.}


\author[1]{D. Gil}[type=editor,
                        auid=000,bioid=1,
                        orcid=0000-0002-2770-4767]
\ead{debora@cvc.uab.cat}

\credit{Conceptualization, Methodology, Software, Validation, Formal analysis, Investigation, Resources, Data Curation, Writing - Original Draft, Writing - Review /& Editing, Visualization, Supervision, Project administration and Funding acquisition}

\affiliation[1]{organization={Universitat Autònoma de Barcelona and Computer Vision Center},
                addressline={Edifici O, Campus UAB}, 
                city={Bellaterra (Cerdanyola)},
                postcode={08193}, 
                state={Barcleona}, 
                country={Spain}}

\author[1]{I. Sanchez}[type=editor,
                        auid=000,bioid=1
                        ]
\ead{isaac.sanchezA@cvc.uab.cat}

\credit{Methodology, Software, Validation, Investigation, Resources, Data Curation}

\author[1]{C. Sanchez}[type=editor,
                        auid=000,bioid=1,
                        orcid=0000-0003-3435-9882]
\cormark[1]
\ead{csanchez@cvc.uab.cat}

\credit{Conceptualization, Methodology, Software, Validation, Formal analysis, Investigation, Resources, Data Curation, Writing - Original Draft, Writing - Review /& Editing, Visualization, Supervision, Project administration and Funding acquisition}

\cortext[cor1]{Corresponding author}


\begin{abstract}
A main barrier for the deployment of AI radiomic systems in clinical routine is their drop in performance under heterogeneous multicentre acquisition protocols. This work presents a performance-oriented framework for quantifying scan parameter sensitivity of radiomic AI models, while identifying clinically significant parameter regions associated with improved cross-dataset robustness. 

Mixed-effects modelling in combination with a hierarchical Pareto-based strategy is used to quantify the influence of acquisition parameters and select critical values associated to low performance of predictive models. We apply our framework to lung cancer diagnosis in CT scans using two independent multicentre datasets (a public dataset and own-collected private data) and several state-of-the-art architectures. To evaluate transferability of results, CT parameters were selected using the private data and validated on the public set. With the selected configurations, validation in low performance conditions has 0.72 ([0.59,0.86], 95\% CI) accuracy and 0.71 ([0.58, 0. 84], 95\% CI) F1Score, while in high performance ones, we obtain 0.88 ([0.7, 1.0], 95\% CI) accuracy with 0.87 ([0.69, 1.0], 95\% CI) F1Score.

\end{abstract}

\begin{graphicalabstract}
\includegraphics[width=1\linewidth]{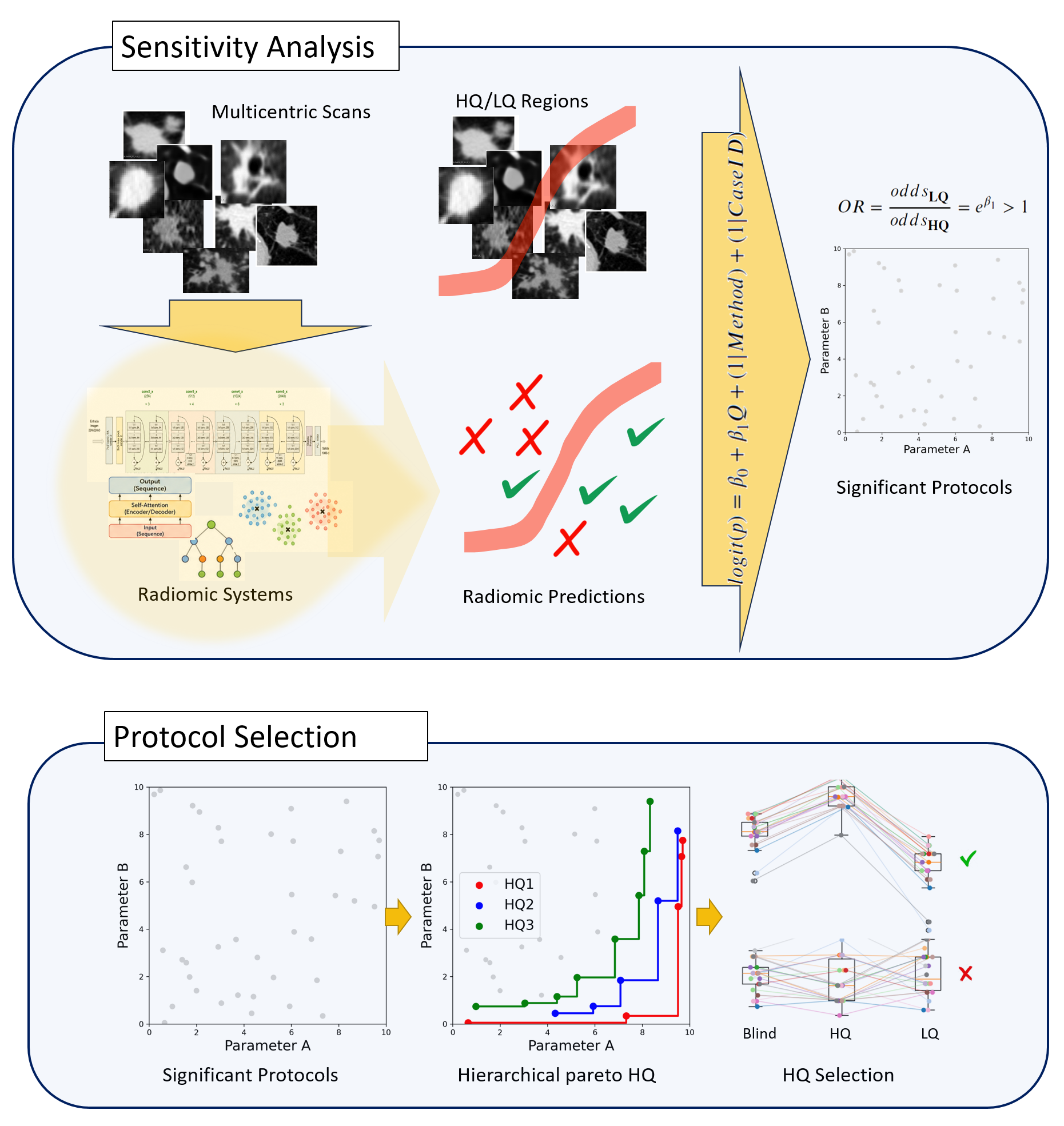}
\end{graphicalabstract}

\begin{highlights}
\item Impact of acquisition variability on performance of models hinders transferability to multi-centre deployment
\item Mixed-effects models allow to quantify AI models sensitivity to acquisition conditions  
\item Hierarchical Pareto strategy identifies parameter regions associated to high-quality performance
\item Validation in CT lung cancer using two independent multicentre datasets (discovery and validation) provides evidence supporting the transferability and potential benefits of the proposed parameter selection strategy
\end{highlights}

\begin{keywords}
AI Models performance \sep
Radiomics Prediction \sep Acquisition Parameter Sensitivity \sep Clinical Acquisition Protocol selection \sep Pareto multi-objective optimization \sep   Generalized Mixed Models \sep 
Framework
\end{keywords}

\maketitle

\section{Introduction}\label{sec_Introduction}

Recent advances in medical image analysis have driven substantial progress in artificial intelligence (AI) models, both handcrafted radiomics and deep learning approaches,  for disease detection and risk stratification. In thoracic oncology, predictive models have demonstrated encouraging performance for lung cancer prediction from computed tomography (CT) scans
\cite{hosseini2023deep}.  
A main obstacle for translation of models into clinical practice is their reduced robustness and generalization when deployed across heterogeneous imaging conditions, which are intrinsic to real-world, multi-centre clinical environments. Such acquisition-induced variations introduce domain shifts in image-derived representations \cite{mackin2015measuring, traverso2018repeatability}, which degrade model performance when applied to unseen acquisition conditions \cite{xu2021effect}. 

To mitigate acquisition-induced variability, several feature-domain harmonization techniques (like ComBat \cite{fortin2018harmonization}) have been adopted. Although harmonization has shown to reduce scanner-related batch effects, its influence on downstream clinical performance remains variable and highly dependent on study design, data composition, and modeling choices \cite{ibrahim2021application, ibrahim2022impact}.  More recent work has explored image-domain normalization using convolutional neural networks or generative adversarial networks \cite{refaee2022ct, yadav2025comparative}. While promising, these approaches typically require large, carefully curated multi-centre datasets and extensive validation to ensure preservation of diagnostic information. In this context, contrastive techniques have been explored as a few-shot method for transforming scans to a common representation space invariant to acquisition parameters \cite{wei2023ctflow}.  

An alternative and comparatively under-explored strategy is to explicitly model acquisition-induced variability within the predictive analysis itself. Statistical models with random effects are particularly well suited to this setting, as they naturally account for hierarchical and clustered data structures arising from multi-scanner, multi-centre cohorts. 
Mixed-effects modeling has previously been used \cite{sun2018radiomics} to include acquisition parameters as factors in regression models for the development of a radiomic signature predicting response to immunotherapy treatments. 
However, its application to quantifying how specific acquisition parameters influence the risk of prediction failure remains limited. Up to our knowledge, there is a lack of validation frameworks for assessing the sensitivity of AI models to scan parameters and adjusting their values for achieving optimal clinical performance. 

In the particular case of CT scans, a substantial body of phantom and patient-based studies has systematically characterized the sensitivity of radiomic features to acquisition-induced variability. Slice thickness and reconstruction kernel consistently emerge as dominant sources of variation, particularly for texture features, whereas tube voltage and pitch typically exert more moderate effects \cite{li2022impact, rinaldi2022reproducibility, foy2020harmonization}. In \cite{ligero2019selection} the stability of radiomic features under different CT acquisition conditions was used as a selection criterion for training radiomic models, showing an increase in clinical performance in lung cancer diagnosis. 
Reproducibility and predictive performance are distinct properties, and feature-level reproducibility does not directly guarantee improved model generalization under heterogeneous acquisition protocols \cite{wennmann2025reproducible, demirciouglu2025rethinking}. 

In this work, we present a framework for performance-oriented statistical analysis of acquisition-induced variability in radiomic systems. We propose a generalized mixed-effects modeling framework to explicitly quantify the influence of relevant acquisition parameters on models' performance, while accounting for subject-level random effects. The significant parameter configurations are further stratified and selected using a multi-objective statistical strategy. To demonstrate the clinical validity of our statistical framework, we have applied it to adjustment of CT scan parameters for lung cancer diagnosis. Several deep learning SoA architectures and radiomic domains are evaluated using two independent datasets: the well-established public benchmark LUNA16  \cite{setio2017validation} and own-collected cases from a multi-centric study RadioLung \cite{data1972_2025}. To assess inference and cross-dataset transferability, acquisition parameters are selected in RadioLung and validated in LUNA16 (henceforth LUNA).

\section{Assessment of Sensitivity to Acquisition Parameters}\label{Sec:SensModels}

A Generalized Linear Mixed Model (GLMM) \cite{MixedMdels} can be formulated as: 

\begin{equation}\label{GLMM}
Y = \beta_0 + \sum_{n} \beta^n_1 x_n +\sum_{m} \beta^m_2 \delta_m +\sum_{n,m} \beta^{n,m}_3  x_n  \delta_m+ (1|\alpha)
\end{equation}
This model estimates the correlation between a dependent variable $Y$ and some explanatory variables, which can be continuous, $x_n$, or categorical, $\delta_m$. In the first case, the model is a regression, while, in the second one, it behaves like a multivariate analysis with groups defined by the factor $\delta_m$. In case of mixed explanatory variables, there is an extra term, $x_n  \delta_m$, modeling across-variable interactions. GLMMs can also incorporate random effects $\alpha \sim N(0,\sigma_{\alpha}^2)$ to account for inter-subject variability and compensate for redundancies due to repeated measurements. 

\subsection{Models for Parameters Adjustment}\label{Sec:ParamOpt}

We use the Odds Ratio (OR) to assess the impact of acquisition conditions on the performance of a predictive method. OR calculates the relationship between a variable and the likelihood of an event occurring. In our case, OR is the risk of error in model predictions for acquisition parameters associated to high (labelled {\bf HQ}) and low (labelled {\bf LQ}) quality performance of models. From now one, we will also refer to scans associated to {\bf HQ} and {\bf LQ} as, respectively, high and low quality scans. 

We estimate OR using a logistic GLMM under a binomial distribution, logit link and fixed factor the quality of scans. Let $f \in \{0,1\}$ be a binary variable indicating case-wise diagnostic failure, with $f=1$ for an incorrect prediction, and $Q\in \{0,1\}$ be a binary factor indicating the quality of the scan, with $Q=0$ for {\bf HQ} cases. Then, the GLMM for the conditional probability $p=P(f=1 | Q)$ given by: 
\begin{equation}\label{LME_Single}
logit(p)=log(\frac{p}{1-p})=log(odds) = \beta_0 + \beta_1 Q 
\end{equation}
\noindent estimates the sensitivity of the predictive method. The exponential $e^{\beta_0}$ estimates odds of model failures in {\bf HQ} scans, while $e^{\beta_1}$ estimates the odds ratio of the risk of failure in {\bf LQ} scans. A significant positive $\beta_1$ indicates that the model is sensitive to scan quality and performs worse in {\bf LQ}:
\[
OR=\frac{odds_{{\bf LQ}}}{odds_{{\bf HQ}}}=e^{\beta_1}>1 \implies odds_{{\bf LQ}} > odds_{{\bf HQ}}
\]
By adding a random effect for the predictive method, we can assess the impact of \textbf{LQ} conditions for methods belonging to the same family of predictive approaches, regardless of the particular sampling used to estimate GLMM models. The formulation of such population-inference GLMM is: 
\begin{equation}\label{LME_Poblational}
logit(p) = \beta_0 + \beta_1 Q +(1 | Method) + (1| CaseID)
\end{equation}
\noindent for $(1 | Method)$, $(1|CaseID)$  random effects modeling, respectively, across-method and inter-patient variability. 

The factor $Q$ grouping scans into \textbf{HQ} and \textbf{LQ} is defined by thresholding the values of the scan parameters. Let each acquisition configuration be represented as a point, $\mathbf{P}$, in the $N_P$-dimensional space of parameters, $\mathbf{P} = (P_1, P_2, \dots, P_{N_P}) \in \mathbb{R}^{N_P}$. We assume that scan quality is an increasing ($\leq$) or decreasing ($\geq$) function of the parameter values depending on the parameter specific direction and define the \textbf{HQ} region as:
\begin{equation}\label{ThQ_Cube}
 Q=Q(\mathbf{Th_Q})= \left\{
\begin{array}{ll}
0 & if \ Th_j [\leq or \geq] P_j, \ \ \forall j \in {1, \dots, N_P}\\
1 & otherwise  \\
\end{array} 
\right .
\end{equation}
The thresholds $\mathbf{Th_Q}=(Th_j)_{j=1}^{N_P}$ can be adapted for optimal performance of a predictive model using $\beta_1$ significance in the multi-objective strategy explained in Section \ref{Sec:ParamOptAlgorithm}. 

\subsection{Models for Analysis of Performance Sensitivity}\label{Sec:ModAssessment}

The GLMM given by (\ref{LME_Poblational}) is  model-agnostic and, thus, it cannot account for specific across-method comparisons or method-parameter interactions. In order to extract model-specific conclusions, we re-formulate (\ref{LME_Poblational}) with the method as fixed effect. Let $\delta_n$, $n=0,\dots N_n$, be the characteristic function of the $n$-th predictor and set the method with the highest performance to be the reference $0$-th predictor. Then, the GLMM formulated as:
\begin{equation}\label{LME_Joint}
logit(p) = \beta_0 + \beta_1 Q + \sum_{n>=1} \beta_2^n \delta_n + \sum_{n>=1} \beta_3^n Q\times\delta_n + (1|CaseID)
\end{equation}
\noindent estimates predictor-specific odds, while accounting for differences among them. Evaluation of (\ref{LME_Joint}) back-transformed to the original scale, estimates the odds for each predictor:
\begin{equation}
\begin{array}{lcl}
odd_{{\bf HQ}_0} = e^{\beta_0} &
; & odd_{{\bf HQ}_n} = e^{\beta_0}e^{\beta_2^n}, \ \ n>0\\
odd_{{\bf LQ}_0} = e^{\beta_0}e^{\beta_1} &
; &  odd_{{\bf LQ}_n} = e^{\beta_0}e^{\beta_1}e^{\beta_2^n}e^{\beta_3^n}, \ \ n>0
\end{array}
\end{equation}
\noindent for $odd_{\textbf{LQ}_n}$, $odd_{\textbf{HQ}_n}$ the n-th predictor odds in, respectively, \textbf{HQ} and \textbf{LQ} conditions. Consequently, the estimated OR comparing methods across scan quality are given by:
\begin{equation}
\begin{array}{ll}
 OR_0=\frac{odd_{{\bf LQ}_0}}{odd_{{\bf HQ}_0}}=e^{\beta_1} & 
 OR_n= \frac{odd_{{\bf LQ}_n}}{odd_{{\bf HQ}_n}}=e^{\beta_1}e^{\beta_3^n} \\ \\
 OR_{\textbf{HQ}_{n,0}}=  \frac{odd_{{\bf HQ}_n}}{odd_{{\bf HQ}_0}}=e^{\beta_2^n}  & 
OR_{\textbf{LQ}_{n,0}}=\frac{odd_{{\bf LQ}_n}}{odd_{{\bf LQ}_0}}=e^{\beta_2^n}e^{\beta_3^n}
\end{array}
\end{equation}
\noindent being $OR_n$, $n=0,\dots, N_n$, the odds ratio for the n-th predictor and $OR_{\textbf{HQ}_{n,0}}$, $OR_{\textbf{LQ}_{n,0}}$ the odds ratio comparing performance of the n-th predictor, $n>0$, to the reference one in \textbf{HQ} and \textbf{LQ} conditions, respectively. Similar to (\ref{LME_Single}), strictly positive and significant $\beta_1$, $\beta_1+\beta_3^n$ indicate sensitivity to the quality condition for, respectively, the reference and the $n$-th ($n>0$) predictors. Regarding across-predictor comparison, n-th predictor is worse in \textbf{HQ} and \textbf{LQ} conditions if $\beta_2^n$, $\beta_2^n+\beta_3^n$ are, respectively, strictly positive and significant.

Performance degradation rate can be also assessed by incorporating continuous variables in the estimation of OR. Let $d_Q$ be the signed distance (negative values for interior points) of $\mathbf{P}$ to $\mathcal{R}(\mathbf{Th_Q})$, then the GLMM formulated as:
\begin{equation}\label{LME_Joint_dQ}
logit(p) = \beta_0 + \beta_1 d_Q + \sum_{n \geq 1} \beta_2^n \delta_n + \sum_{n\geq 1} \beta_3^n d_Q\times\delta_n + (1|CaseID)
 \end{equation}
\noindent approximates OR as a function of parameters distance to the optimal values accounting for differences across the tested predictor methods. The expected odds, $odd_n$, $n\geq 0$, and corresponding ORs as function of $d_Q$ are given by:
\begin{equation}
\begin{array}{lcl}
odd_{0} = e^{\beta_0}e^{\beta_1 d_Q}&
; & odd_{n} = e^{\beta_0+\beta_2^n}e^{(\beta_1+\beta_3^n)d_Q}, \ \ n>0 \\
OR_{0} = e^{\beta_1 d_Q}&
; & OR_{n} = e^{(\beta_1+\beta_3^n)d_Q}, \ \ n>0
\end{array}
\end{equation}
 
The exponential $e^{\beta_1}$ estimates the rate of performance decay in best method odds per unit increase in distance to the \textbf{HQ} region, while $e^{\beta_3^n}$ compensates for deviations in decay for the other methods. The model (\ref{LME_Joint_dQ}) can support inference to the whole family of sampled predictive methods by adding the method as random effect:
\begin{equation}\label{LME_Joint_dQ_Population}
logit(p) = \beta_0 + \beta_1 d_Q + (1|Method) + (1|CaseID)
\end{equation}
We would like to note that formulations with fixed effects should be used for comparison of sensitivity among a specific set of predictive methods, while random effects are intended to leverage conclusions and inference to a family of AI methods. In this last case, a large enough number of representative AI methods have to be evaluated and considered to estimate models. 

\section{Multi-objective Selection of Acquisition Parameters}\label{Sec:ParamOptAlgorithm}

The collection of optimal thresholds is obtained by a multi-objective selection of a grid search of each parameter dimension range given by the minimum and maximum values allowed by the scan. 


Let $(\mathbf{Th_Q}^s)_s$ be the collection of threshold values such that $\beta_1$ estimated by (\ref{LME_Poblational}) is significant.  Given that flexible acquisition parameters favour multicentre deployments, the optimal thresholds should define maximal regions in the parametric space $\mathbb{R}^{N_P}$. Since such regions are defined by an inequality in each dimension, it follows that coordinates of the optimal thresholds should dominate as many significant threshold values as possible. That is, they are the Pareto solution \cite{coello2007evolutionary} to the multi-objective optimization problem given by the coordinates of the significant threshold values. 

Pareto optimization was selected because acquisition protocol selection is inherently a multi-objective problem. The quality region is simultaneously determined by several acquisition parameters and no clinically justified scalar utility function exists to combine them into a single optimization criterion. Weighted approaches would require subjective weighting factors between parameters. Pareto dominance avoids this limitation and provides a family of equivalent non-dominated solutions that can subsequently be adapted to scanner-specific clinical constraints. Rather than restricting the analysis to a single Pareto front, we propose an iterative hierarchical Pareto decomposition strategy to provide with a collection of thresholding fronts with different quality configurations. At each iteration,
all non-dominated configurations are identified and assigned to a high quality layer denoted as $\mathrm{HQ}_\ell$. These configurations are then removed from the candidate set, and the process is repeated until no configurations remain. This procedure yields an ordered set of Pareto layers, $\{\mathrm{HQ}_1, \mathrm{HQ}_2, \dots, \mathrm{HQ}_\ell\}$, with $\mathrm{HQ}_1$ defining globally optimal trade-offs and higher layers corresponding to progressively relaxed multi-objective compromises. Each configuration $\mathbf{Th_Q}^s \in \mathrm{HQ}_\ell$ induces an admissible region in the parameter space, $\mathcal{R}(\mathbf{Th_Q}^s)$  defined by (\ref{ThQ_Cube}).
The union of all regions of a Pareto layer $\mathrm{HQ}_\ell$ represents the full set of parameter configurations that are multi-objectively equivalent to quality level
$\mathrm{HQ}_\ell$, providing a set-valued description of acceptable solutions rather than a single operating point.

Some Pareto layers might include extreme thresholds defining too restrictive regions that still might either exclude good configurations or include low quality ones, which size was not enough to reduce effect significance. From the clinical translational point of view, this can be undesirable for full deployment. Besides, we consider that in order that our selection process gives some clinical benefit, splitting should guarantee that metrics under \textbf{HQ} and \textbf{LQ} conditions are, respectively better and lower than the ones obtained on a sampling of the whole population including any quality condition (labelled \textbf{BLIND} since the quality condition is unknown). Under this consideration, Pareto thresholds having the largest average differences \textbf{LQ}-\textbf{BLIND},  \textbf{HQ}-\textbf{BLIND} and \textbf{LQ}-\textbf{HQ} are the ones that are finally selected. 
Prioritization of the 3 differences ordering for decision making might vary depending on the specific clinical problem. However, a good decision rule is to select the configuration having the largest difference in \textbf{HQ}-\textbf{BLIND} and use \textbf{LQ}-\textbf{BLIND} for ties. 



\begin{figure}
    \centering
    \includegraphics[width=1\linewidth]{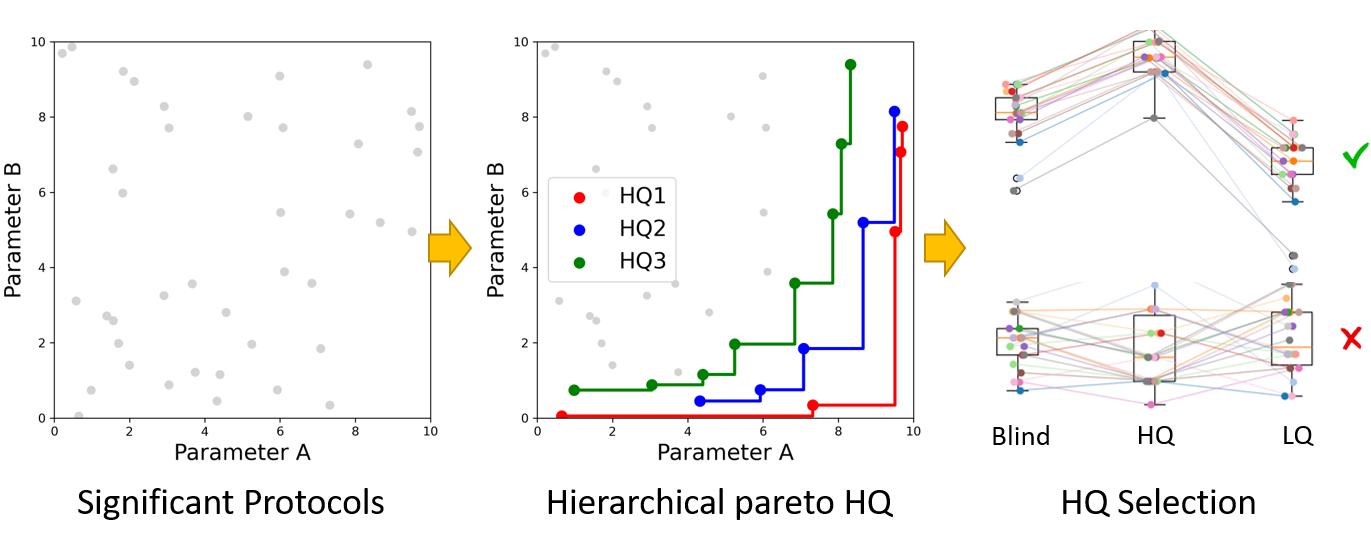}
    \caption{Scheme of the main steps in the multi-objective selection of acquisition parameters.}
    \label{fig:esqHQMetod}
\end{figure}

Figure \ref{fig:esqHQMetod} shows the principal steps of the framework in a 2-dimensional search space with increasing quality in ParameterA (x-axis) and decreasing quality in ParameterB (y-axis). The most-left plot shows in gray dots the initial collection of clinical protocols selected based on GLMM significance of a discretization of the 2-dimensional search space. The hierarchical Pareto front for the stratification in $ \mathrm{HQ}_\ell$ layers is shown in the middle plot. We show 3 layers of decreasing quality with their sets of parameter combinations in solid dots. Finally, the boxplots of the most-right plot shows 2 examples of the final selection of best clinical protocols based on statistical tests comparing performance between \textbf{LQ}, \textbf{HQ} and \textbf{BLIND} metrics. Upper boxes show an example of valid configuration having significantly increasing performance in \textbf{LQ}, \textbf{BLIND} and \textbf{HQ} scans, in comparison to the lower boxes which show equal performance across the 3 conditions. 

\section{Application to Lung Cancer Diagnosis in CT Scans}
We have applied our framework to assess the impact that CT scan parameters have on methods for lung cancer diagnosis. In particular, we focus on 3 main parameters: Xray tube (labelled $XRay$), Spiral pitch (labelled $Spiral$), and Slice thickness (labelled $SliceTh$). According to radiological guidelines \cite{xu2021effect}, these parameters define the following multi-objective dominance:
\[ \mathcal{R}(\mathbf{Th_Q})=\{Xray\geq Th_{Xray},  Spiral \leq Th_{Spiral}, Slice \leq Th_{SliceTh}\}
\]


We have compared two representation spaces for pulmonary nodule analysis: a deep textural representation derived from texture volumes \cite{torres2022intelligent,baeza2024radiomics}, and a conventional intensity-based representation. In both cases, the representations are processed using pre-trained convolutional neural networks and classified using a fully connected (FC) networks. 

The Intensity Representation space uses directly intensity values of the lesion as input to pre-trained extractors. Each volume of interest (VOI) of a pulmonary nodule is decomposed into axial slices as inputs to the feature extractor. The embedded slices define the Intensity Representation. In the Radiomic Representation space the VOI is first described using three-dimensional gray-level co-occurrence matrix (GLCM) descriptors \cite{lofstedt2019gray} as described in \cite{torres2022intelligent}. Each GLCM-derived volume is decomposed in axial slice by slice as inputs to the feature extractor. The deep features extracted from all GLCM volumes are concatenated to define the Radiomic Representation.

Regarding backbone extractors, we have used several architectures, including a wide range of CNNs (VGG16 \cite{simonyan2014very}, VGG19,  MobileNetV2 \cite{sandler2018mobilenetv2}, MobileNetV3l, DenseNet-169 \cite{he2016deep}, ResNet50, ResNet-152, ResNet18 \cite{nagpal2022comparative}, EfficientNet-B7 \cite{tan2019efficientnet}, EfficientNetV2l, EfficientNetV2s), visual transformers (ViTB16 \cite{wu2020visual}) and hybrid models (SwinB \cite{zhang2022transformer}, ConvNextB \cite{todi2023convnext}). 
 
Features extracted for each representation domain and network are the input to a 1 layer FC network. Back-bone encoders were freeze during training and only the classification head was trained for slice-level malignancy classification. To mitigate class imbalance a weighted cross-entropy loss with weights given by inverse frequencies was used. Slice predicted probabilities were pooled using nodule-level averages to obtain a malignancy score at nodule level.


\subsection{Datasets}

\begin{figure}
    \centering
    \begin{subfigure}{0.9\textwidth}
        \centering
        \includegraphics[width=\textwidth]{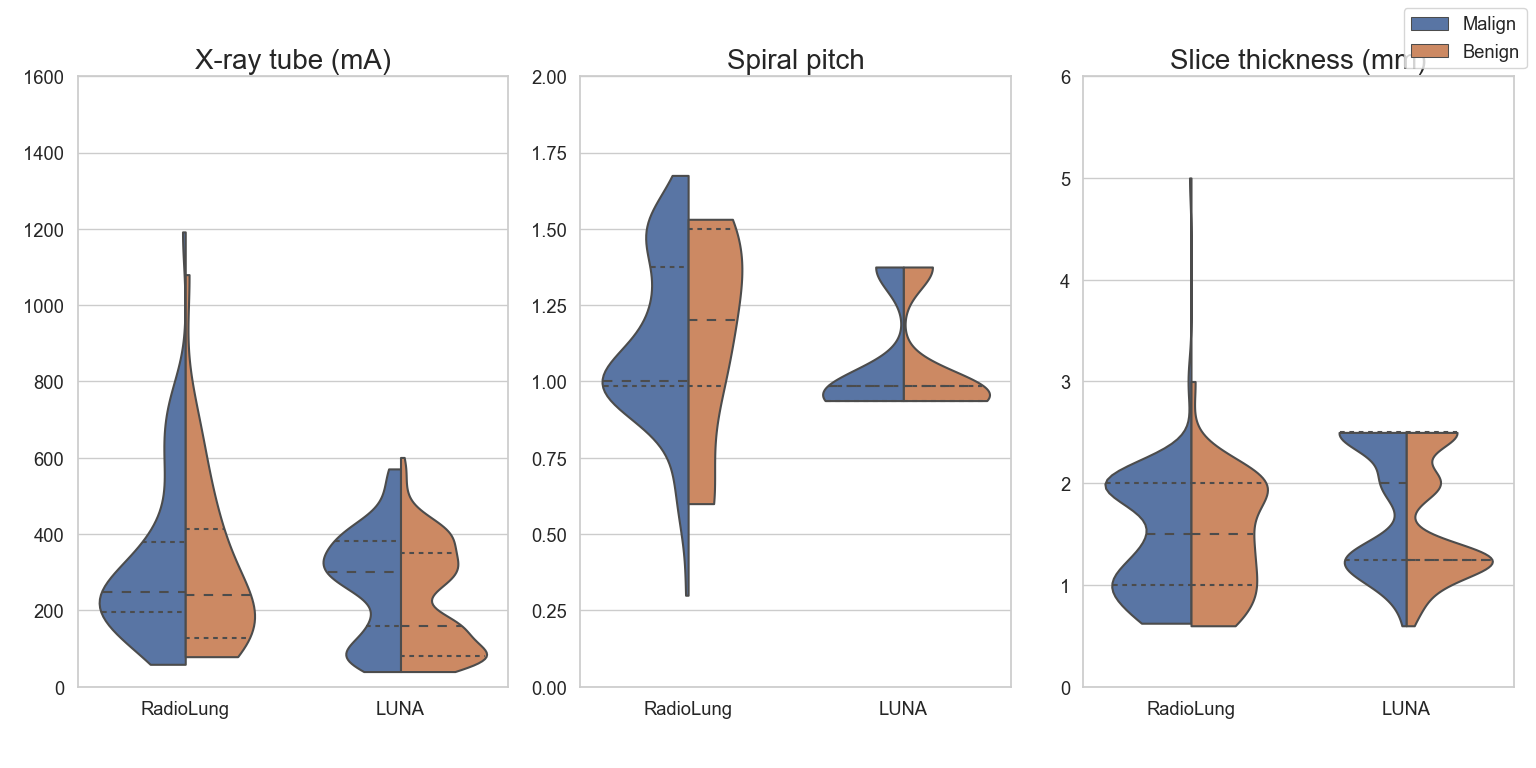}
        \caption{RadioLung and LUNA datasets}
        \label{fig:regregion}
    \end{subfigure}
    \\
    \begin{subfigure}{0.9\textwidth}
        \centering
        \includegraphics[width=\textwidth]{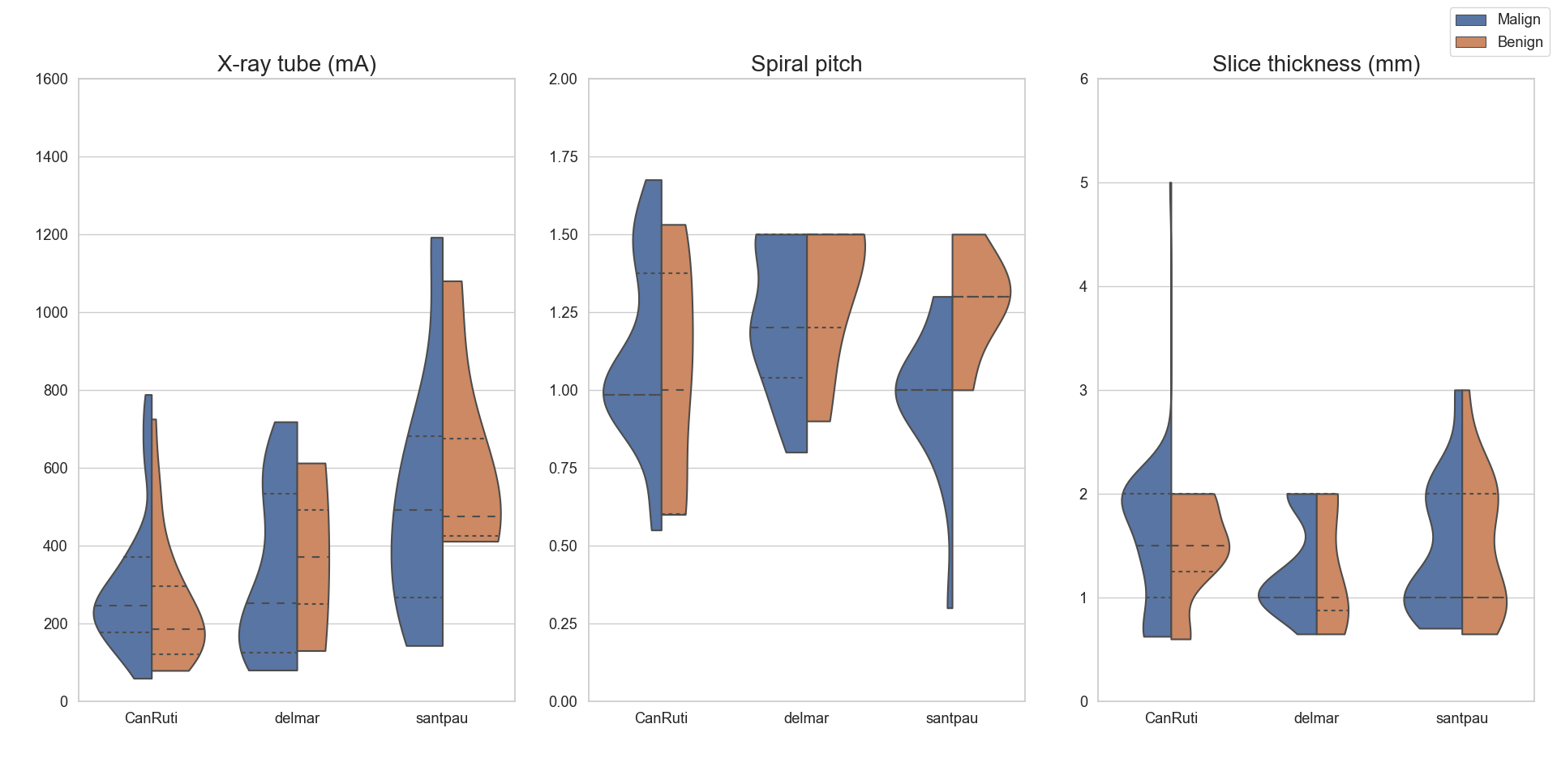}
        \caption{RadioLung center specific}
        \label{fig:iregion}
    \end{subfigure}
    \caption{Violin plots (cut=0) showing distribution of CT acquisition parameters across samples of LUNA and RadioLung datasets (a) and RadioLung center specific (b). }
    \label{fig:DistParam}
\end{figure}

We have used two independent multicenter datasets: the public benchmark LUNA \cite{setio2017validation} and the own-collected RadioLung  \cite{data1972_2025}. LUNA cases were filtered to have at least three radiological diagnosis. The diagnostic label was assigned from the average malignancy score (1 to 5) provided by radiologists as: Benign for average score between 1 to 2.5; Malign for average score between 3.5 to 5; Indeterminate for average score between 2.5 and 3.5. These last cases were also excluded. 
A total of 987 nodules (644 benign and 343 malign) were selected and completed with their DICOM metadata. The RadioLung dataset (see \cite{baeza2024radiomics,data1972_2025} for more details) of small nodules contains nodules gathered from 3 Spanish centers (Hosp. Germans Tries i Pujol, Hosp. Sant Pau, Hosp. Del Mar) with scans acquired with 7 different devices. A total number of 201 solitary nodules were scanned and diagnosed through biopsy (51 identified as benign). Pre-processing details (including VOI construction, voxel resampling and HU preprocessing) can be found in \cite{torres2022intelligent,baeza2024radiomics}. 

Both datasets had scans with missing metadata related to acquisition: 807 nodules (278 malign, 529 benign) for LUNA and, 93 nodules (69 malign, 24 benign) for RadioLung. These scans were excluded for GLMMs analysis. The distribution of the acquisition parameters for the remaining scans is shown in Figure \ref{fig:DistParam}. Figure\ref{fig:DistParam} (a) shows the distribution for each dataset and fig. \ref{fig:DistParam}(b) center-specific distribution for RadioLung disaggregated by malignancy. For both datasets, we do not observe any bias or pattern in parameters that could be associated to malignancy and, thus, introduce shortcut learning. Even if LUNA is a larger dataset, its parameter distribution is less heterogeneous and covers a narrower range of parameters. Furthermore, for RadioLung we do not observe any center-specific pattern in parameter distribution prone to introduce center-biases or shortcut learnings. This supports the use of RadioLung as a discovery cohort to explore optimal configurations and block LUNA as independent validation dataset. 



\subsection{Experimental Design}
Radiolung dataset has been considered as a discovery cohort for parameter adjustment and selection of best performing models. LUNA is used as the validation dataset to assess transferability of the selection of high quality acquisition parameters and good performing models. Specifically, our experiments test the following:

\begin{enumerate}
    \item {\bf Parameters Adjustment.} Parameter ranges were selected using RadioLung cases excluding missing values. The parametric space is defined by the ranges of CT scans used in radiomic studies: $SliceTh \in [0.5,5 ]$, $Spiral \in [0.3,1.5]$, $XRay \in [50,800]$. This space was discretized into adaptive bins to ensure a minimum number of samples per-bin for the grid search method described in Section \ref{Sec:ParamOptAlgorithm}. Bins breaks for each parameter were $[1,1.25,1.5,2,2.5]$, $[0.6,1,1.2,1.3,1.4]$, $[100,150,200,300,350,400,450,500]$ for, respectively, $Slice$, $Spiral$ and $XRay$. 
    
    \item {\bf Performance Sensitivity.} The impact of Pareto optimal parameters was evaluated in both datasets using the model-specific formulations of Section \ref{Sec:ModAssessment}. Independent models were adjusted for each set. Different fixed-effects models were adjusted for each dataset to assess method-specific sensitivities. The overall sensitivity was assessed using a population continuous model (\ref{LME_Joint_dQ_Population}). Distances were computed using the maximum norm and $XRay$ units have been transformed to one tenth of an ampere by dividing by 100 in order to have comparable scales across parameters.  

    \item {\bf Clinical Impact.} In this last experiment we validated the selected parameters in terms of an increase in diagnostic performance of models. Performance was assessed using sensitivity, specificity, F1-score and weighted accuracy. Bootstrap confidence intervals were used to report variability in metrics in {\bf LQ}, {\bf HQ} and {\bf BLIND} groups. Assessment of the increase in clinical diagnostic metrics has been evaluated only on the validation dataset LUNA. 
\end{enumerate}


Independent models were trained for each dataset using a 10-fold stratified split at patient level to avoid data leakage and maintain similar nodule distribution across folds. All available data (including missing metadata scans) was used for training and computing out-of-fold (oof) predictions. For RadioLung, we used 2 different k-fold partitions, one for parameter adjustment and the other one for assessment of performance sensitivity and model selection. All models were trained for 50 epochs with a learning rate of 0.001 and Adam optimizer. No hyperparameter tuning was conducted. 

Independent GLMMs were adjusted for each dataset using oof predictions of the cases with scans without missing metadata. Models achieving the highest oof accuracy on the full data set were selected as the reference one. In the case of RadioLung, cases with scan missing data define the \textbf{BLIND} population pool to compute metrics for the final selection of parameters. The p-values of the $\beta_1$ used for parameter selection were corrected using FDR/Benjamini-Hochberg to account for the multiple parameter configurations compared. 

Statistical analysis was conducted using R version 4.3.3. GLMMs were estimated with glmer() function of the lme4 package using maximum likelihood (Laplace Approximation) estimation and Bound Optimization BY Quadratic Approximation optimization algorithm \cite{powell2009bobyqa} to help convergence. Post-hoc multiple comparisons tests for the significance of odds ratio were performed 
using estimated marginal means (EMMeans), applying Tukey's method for p-value adjustment. Tests were performed on the log odds ratio scale and estimated OR were back transformed to the original scale for better interpretation. For all statistical analysis, a p-value $<0.05$ was considered significant.


\begin{table}
\centering
\caption{Performance (accuracy) comparisons for Pareto-front regions}
\label{tab:RadioLungParetoFrontAnovaCompact}
\begin{tabular}{lclcr}
\toprule
\textbf{Region} & \textbf{Samples group} & \textbf{Groups} & \textbf{meandiff} & \textbf{CI} \\
\midrule

$\mathrm{HQ}_1$: [200, 1.0, 1.25]
 & 93/27 & BLIND/HQ & 0.15 & [0.10, 0.20]  \\
  & 93/81 & BLIND/LQ & -0.005 & [-0.05, 0.04]  \\
  & 27/81 & HQ/LQ  & -0.16 & [-0.20, -0.11]  \\
\midrule

$\mathrm{HQ}_2$: [300, 1.2, 1.0]
 & 93/16 & BLIND/HQ & 0.17 & [0.12, 0.21   ]  \\
 & 93/92 & BLIND/LQ & 0.01 & [-0.03, 0.05]  \\
 & 16/92 & HQ/LQ  & -0.16 & [-0.21, -0.12]  \\
\midrule

$\mathrm{HQ}_3$: [200, 1.2, 1.0]
 & 93/29 & BLIND/HQ & 0.19 & [0.14, 0.23]  \\
  & 93/79 & BLIND/LQ & -0.02 & [-0.07, 0.02]  \\
  & 29/79 & HQ/LQ  & -0.21 & [-0.26, -0.17]  \\
\midrule

$\mathrm{HQ}_3$: [300, 1.3, 1.0]
 & 93/19 & BLIND/HQ & 0.16 & [0.12, 0.20]  \\
 & 93/89 & BLIND/LQ & 0.006 & [-0.04, 0.05]  \\
 & 19/89 & HQ/LQ  & -0.15 & [-0.20, -0.11] \\
\midrule

$\mathrm{HQ}_4$: [300, 1.5, 1.0]
&93/24 & BLIND/HQ & 0.15 & [0.11, 0.19]  \\
 & 93/84 & BLIND/LQ & -0.01 & [-0.04, 0.04]  \\
  & 24/84 & HQ/LQ  & -0.15 & [-0.20, -0.11]  \\
\midrule

$\mathrm{HQ}_5$: [200, 1.5, 1.0]
& 93/39 & BLIND/HQ & 0.17 & [0.13, 0.21]  \\
  & 93/69 & BLIND/LQ & -0.05 & [-0.09, -0.01]  \\
  & 39/69 & HQ/LQ  & -0.21 & [-0.26, -0.17]  \\
\midrule

$\mathrm{HQ}_5$: [300, 1.5, 1.25]
  & 93/28 & BLIND/HQ & 0.15 & [0.11, 0.19]  \\
 & 93/80 & BLIND/LQ & -0.01 & [-0.05, 0.04]  \\
 & 28/80 & HQ/LQ  & -0.16 & [-0.20, -0.12]  \\
\midrule

$\mathrm{HQ}_6$: [100, 1.5, 1.0]
  & 93/58 & BLIND/HQ & 0.13 & [0.09, 0.17]  \\
  & 93/50 & BLIND/LQ & -0.09 & [-0.16, -0.05]  \\
  & 58/50 & HQ/LQ  & -0.22 & [-0.26, -0.18]  \\
\midrule

$\mathrm{HQ}_6$: [200, 1.5, 1.25]
 & 93/44 & BLIND/HQ & 0.17 & [0.12, 0.21]  \\
& 93/64 & BLIND/LQ & -0.06 & [-0.10, -0.02]  \\
 & 44/64 & HQ/LQ  & -0.22 & [-0.27, -0.18]  \\
\midrule

$\mathrm{HQ}_7$: [100, 1.5, 1.25]
  & 93/67 & BLIND/HQ & 0.13 & [0.09, 0.17]  \\
& 93/41 & BLIND/LQ & -0.14 & [-0.17, -0.10]  \\
 & 67/41 & HQ/LQ  & -0.28 & [-0.31, -0.24]  \\
\bottomrule
\end{tabular}
\end{table}

\section{Results}

\subsection{Parameters Adjustment}

\begin{figure}
    \centering
    \includegraphics[width=0.9\linewidth]{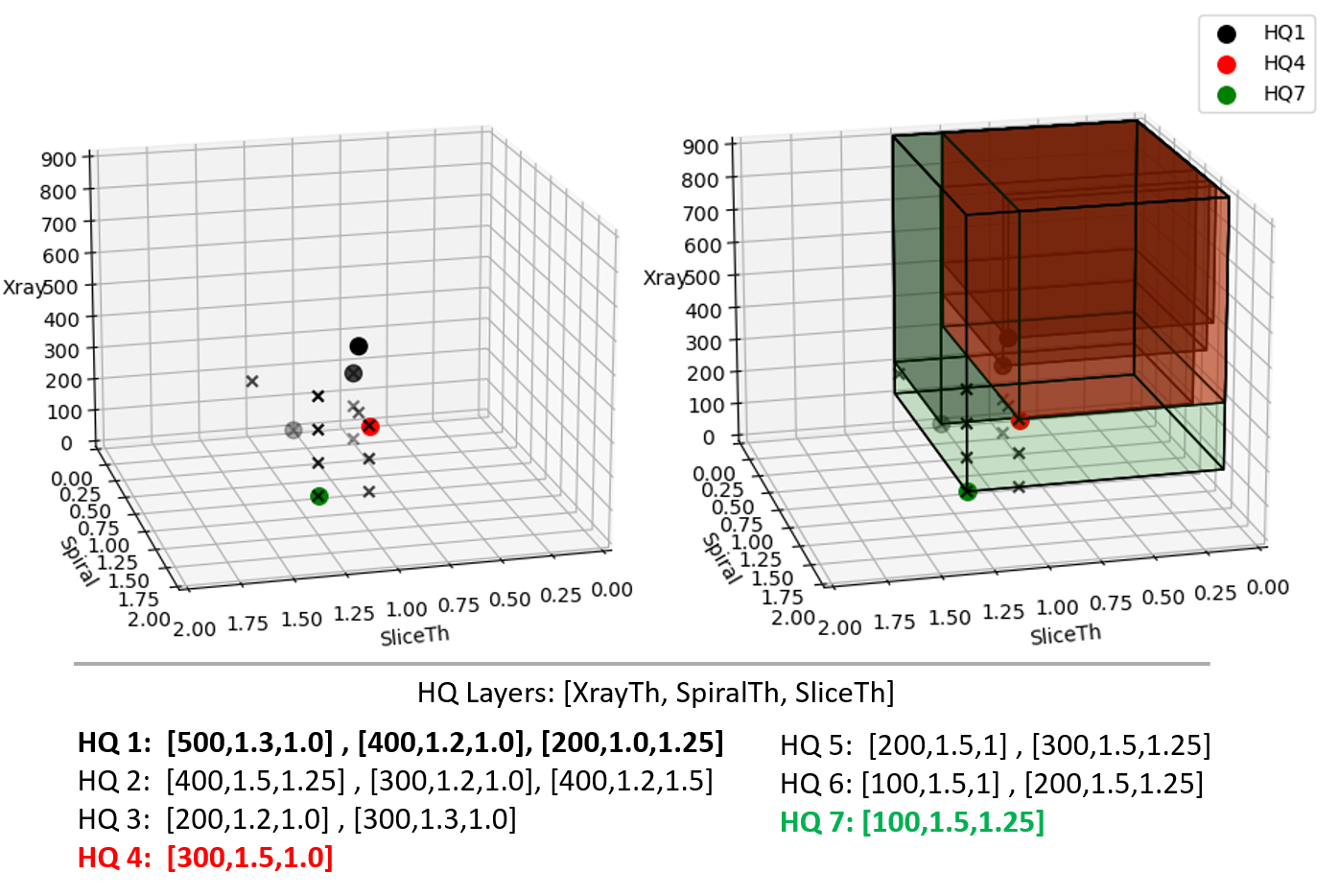}
    \caption{Acquisition parameter multi-objective selection procedure applied to discovery RadioLung dataset}
    \label{fig:esqHQDef}
\end{figure}

Figure \ref{fig:esqHQDef} illustrates our multi-objective selection applied to the RadioLung set. The left plot shows all significant configurations in black crosses and configurations belonging to 3 Pareto layers with progressive relaxed level of multi‑objective optimality in coloured dots: the top $\mathrm{HQ}_1$ in black, the middle $\mathrm{HQ}_4$ in red and the lower $\mathrm{HQ}_{7}$ in green. The right plot shows the three‑dimensional admissible regions bounded by problem‑specific inequalities and global parameter limits defined by the 3 layers. Each semi‑transparent cuboid corresponds to the dominance region of a Pareto layer, and their overlap illustrates the flexibility and redundancy within the same multi‑objective quality level. The thresholds of all Pareto layers are given in the bottom text with hierarchical multi-objective Pareto-based strategy code and interactive visualization of these pareto layers. \footnote{Available in: \textit{https://github.com/IAM-CVC/OptimizeAcquisitionParameters}.}

Table \ref{tab:RadioLungParetoFrontAnovaCompact} summarizes the pair-wise comparison of accuracy under the different scan qualities for the parameter configurations selected by the Pareto front. Accuracy was computed using patient-level predictions pooled for all predictive models. The table reports sample size, mean differences (positive values indicating higher accuracy of the 2n group) and 95\% bootstrap confidence intervals (CIs). Since in lung cancer screening, low dose scans are recommended we only show results for configurations with $XRay \leq 300$. 


As expected, the difference {\bf LQ}-{\bf HQ} is always negative. The configurations that have  positive and negative intervals for, respectively,  {\bf HQ}-{\bf BLIND} and {\bf LQ}-{\bf BLIND} are: [200,1.5,1], [200,1.5,1.25],[100,1.5,1] and [100,1.5,1.25]. These configurations have also the largest difference between {\bf HQ} and {\bf LQ}. The prioritization given by decreasing differences in {\bf HQ}-{\bf BLIND} and {\bf LQ}-{\bf BLIND} sorts these parameters as: 1) [200,1.5,1.25] (0.17 for {\bf HQ}-{\bf BLIND}, -0.06 for {\bf LQ}-{\bf BLIND}); 2) [200,1.5,1.0] (0.17 for {\bf HQ}-{\bf BLIND}, -0.05 for {\bf LQ}-{\bf BLIND}); 3) [100,1.5,1.25] (0.13 for {\bf HQ}-{\bf BLIND}, -0.14 for {\bf LQ}-{\bf BLIND}) and 4) [100,1.5,1.0] (0.13 for {\bf HQ}-{\bf BLIND}, -0.09 for {\bf LQ}-{\bf BLIND}). We select [200,1.5,1.25] for its validation in LUNA. 


\subsection{Performance Sensitivity}\label{Sec:Exp2}

\begin{table*}
\centering
\caption{Fixed-effects models grouped by architecture and representation for the LUNA validation and RadioLung discovery datasets.}
\label{tab:RegressionModels_ByArch_Vertical_Final}
\begin{tabular}{lcccccccc}
\toprule
\textbf{Architecture}
& \multicolumn{4}{c}{\textbf{GLCM3D}}
& \multicolumn{4}{c}{\textbf{Intensity}} \\
\cmidrule(lr){2-5}\cmidrule(lr){6-9}
& $OR_n$ & pval & $OR_{\mathbf{HQ}_{n,0}}$ & pval
& $OR_n$ & pval & $OR_{\mathbf{HQ}_{n,0}}$ & pval \\
\midrule

\multicolumn{9}{l}{\textbf{LUNA Validation dataset}} \\
\midrule

Swin-B
& 4.8 & 0.02 & -- & --
& 2.1 & 0.26 & 1.8 & 0.27 \\

ConvNextB
& 5.7 & 0.01 & 1.8 & 0.27
& 4.2 & 0.04 & 1.3 & 0.58 \\

DenseNet169
& 5.5 & 0.01 & 2.4 & 0.09
& 2.5 & 0.18 & 1.8 & 0.27 \\

EfficientNetB7
& 6.4 & 0.01 & 2.4 & 0.09
& 1.2 & 0.78 & 3.1 & 0.03 \\

EfficientNetV2-L
& 5.2 & 0.02 & 1.2 & 0.78
& 0.8 & 0.64 & 4.2 & $<$0.01 \\

EfficientNetV2-S
& 2.3 & 0.21 & 2.4 & 0.09
& 1.1 & 0.94 & 3.6 & 0.01 \\

MobileNetV2
& 4.6 & 0.03 & 1.8 & 0.27
& 2.1 & 0.26 & 1.8 & 0.27 \\

MobileNetV3-L
& 5.0 & 0.02 & 2.0 & 0.17
& 4.5 & 0.03 & 1.5 & 0.41 \\

ResNet152
& 3.3 & 0.08 & 3.6 & 0.01
& 2.2 & 0.25 & 2.7 & 0.05 \\

ResNet18
& 5.6 & 0.01 & 1.3 & 0.58
& 2.0 & 0.30 & 2.7 & 0.05 \\

ResNet50
& 7.0 & $<$0.01 & 2.4 & 0.09
& 1.5 & 0.54 & 3.1 & 0.03 \\

VGG16
& 7.6 & $<$0.01 & 1.3 & 0.58
& 1.7 & 0.43 & 1.8 & 0.27 \\

VGG19
& 3.7 & 0.06 & 2.4 & 0.09
& 1.5 & 0.53 & 3.6 & 0.01 \\

ViT-B/16
& 8.4 & $<$0.01 & 1.5 & 0.41
& 4.6 & 0.03 & 1.8 & 0.27 \\

\midrule
\multicolumn{9}{l}{\textbf{RadioLung discovery dataset}} \\
\midrule

ViT-B/16
& 20 & $<$0.01 & -- & --
& 3.9 & 0.08 & 14 & $<$0.01 \\

ConvNextB
& 74 & $<$0.01 & 0.37 & 0.49
& 17 & $<$0.01 & 3.6 & 0.22 \\

DenseNet169
& 12 & 0.01 & 3.0 & 0.30
& 31 & $<$0.01 & 1.5 & 0.72 \\

EfficientNetB7
& 2.8 & 0.18 & 16 & $<$0.01
& 3.9 & 0.07 & 25 & $<$0.01 \\

EfficientNetV2-L
& 12 & $<$0.01 & 4.4 & 0.15
& 3.4 & 0.11 & 14 & $<$0.01 \\

EfficientNetV2-S
& 4.4 & 0.06 & 10 & 0.02
& 4.0 & 0.07 & 17 & $<$0.01 \\

MobileNetV2
& 5.3 & 0.05 & 4.4 & 0.15
& 12 & $<$0.01 & 4.9 & 0.12 \\

MobileNetV3-L
& 64 & $<$0.01 & 0.37 & 0.49
& 5.6 & 0.04 & 4.9 & 0.12 \\

ResNet152
& 5.3 & 0.05 & 4.4 & 0.15
& 6.1 & 0.02 & 8.6 & 0.03 \\

ResNet18
& 4.5 & 0.07 & 6.1 & 0.08
& 8.3 & 0.01 & 4.9 & 0.12 \\

ResNet50
& 12 & 0.01 & 3.0 & 0.30
& 45 & $<$0.01 & 1.5 & 0.72 \\

Swin-B
& 74 & $<$0.01 & 0.37 & 0.49
& 2.6 & 0.23 & 11 & 0.02 \\

VGG16
& 84 & $<$0.01 & 0.37 & 0.49
& 10 & 0.01 & 3.6 & 0.22 \\

VGG19
& 20 & $<$0.01 & 1.0 & 1.00
& 19 & $<$0.01 & 1.5 & 0.72 \\

\bottomrule
\end{tabular}
\end{table*}

\begin{figure}
    \centering
    \begin{subfigure}{0.4\textwidth}
        \centering
        \includegraphics[width=\textwidth]{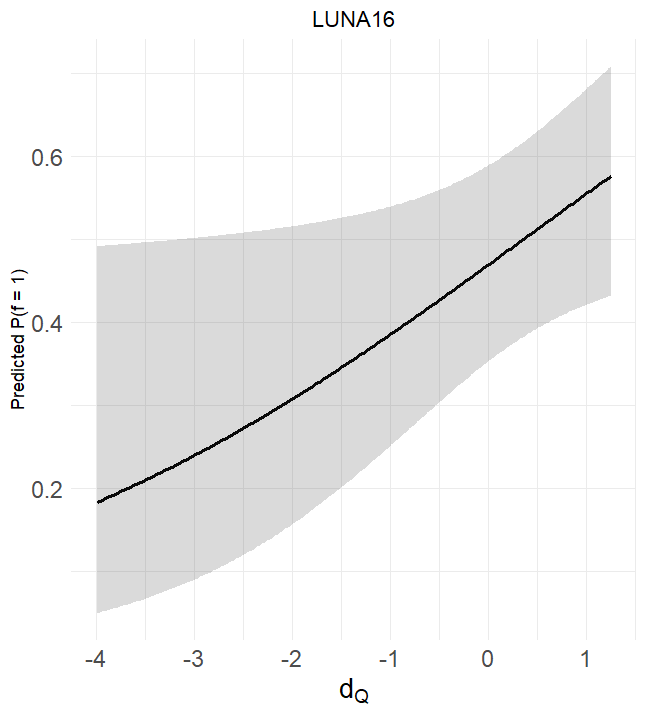}
        \caption{}
    \end{subfigure}
    \begin{subfigure}{0.4\textwidth}
        \centering
        \includegraphics[width=\textwidth]{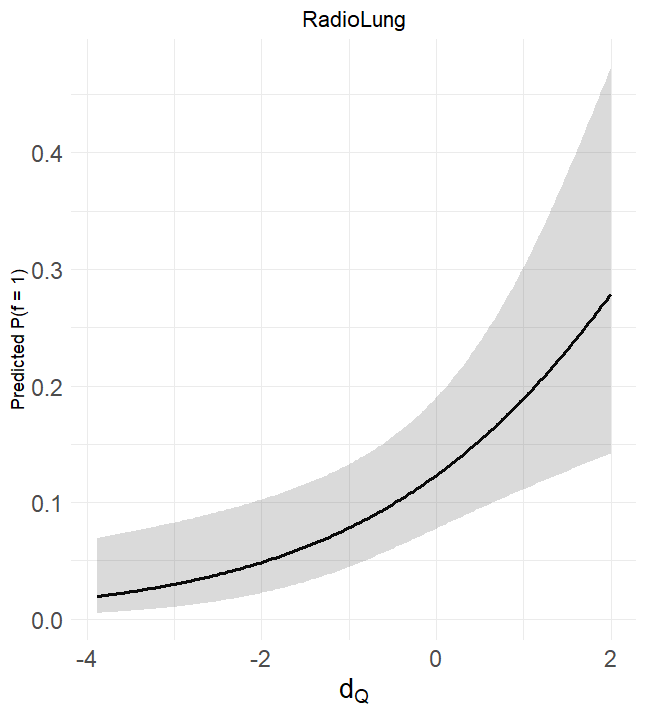}
        \caption{}
    \end{subfigure}
    \caption{Performance decay for LUNA, (a), and RadioLung, (b), datasets. }
    \label{fig:DistanceModels}
\end{figure}

Table \ref{tab:RegressionModels_ByArch_Vertical_Final} reports the fixed-effect GLMMs (\ref{LME_Joint}) adjusted for each dataset using oof predictions. For the discovery RadioLung, models were re-trained using a different 10-fold split. For each GLMM, we report $OR_n$ assessing sensitivity of each method to \textbf{LQ}, $OR_{\textbf{HQ}_{n,0}}$ comparing performance to the reference in \textbf{HQ} and the respective p values for OR significance estimated for each effect. The reference models are GLCM3DSwinB and GLCM3DViTB16 for, respectively, LUNA and RadioLung. 

For, both, discovery RadioLung and validation LUNA performance under \textbf{LQ} is lower ($OR_n>1$) for all AI models. Specifically, the performance drop observed in LUNA supports transportability of the proposed statistical multi-objective parameter selection. Comparison of AI methods in \textbf{HQ} conditions ($OR_{\textbf{HQ}_{n,0}}$), indicates that there are more worse performers in intensity representation space. There are 13 methods with non-significant $OR_{\textbf{HQ}_{n,0}}$ in the discovery set RadioLung. From these methods, 11 have also non significant $OR_{\textbf{HQ}_{n,0}}$ in LUNA. This supports transportability of the method selection given by GLMM analysis on RadioLung. 


The distance models estimated for each dataset are:
\begin{eqnarray}
\label{LUNARadiolung_dQ_Population}
logit(f_{LUNA}) = -0.12 + 0.34 d_Q + (1|Method) + (1|CaseID)
\\
logit(f_{RadioLung}) = -1.96 + 0.51 d_Q + (1|Method) + (1|CaseID) \nonumber
\end{eqnarray}
\noindent The slope $\beta_1$  is significant in both cases with p-values of $0.0387$ and $0.00281$ for, respectively, LUNA and RadioLung. As expected in view of the $OR_n$ ranges estimated in the fixed-effects GLCMs, the increase in error is steeper for the RadioLung model. 
Figure \ref{fig:DistanceModels} plots the regression models back transformed to the original scale with 95\% confidence intervals for failure probability estimation in gray shadowed bands. Negative $d_Q$ indicate scans of very high quality, while positive values are for low quality scans. For each dataset, $d_Q$ axis covers the ranges of each dataset scan parameter distribution.

\subsection{Clinical Impact}\label{Sec:ClinImpact}

Assessment of the increase in clinical diagnostic metrics in LUNA has been evaluated for the methods not having significant performance drop in \textbf{HQ} conditions ($OR_{\textbf{HQ}_{n,0}}$) on RadioLung. For the sake of compactness, we only show results for the 5 methods having the smaller $OR_{\textbf{HQ}_{n,0}}$ in the discovery RadioLung set. Remarkably, LUNA reference model in Table \ref{tab:RegressionModels_ByArch_Vertical_Final} (GLCM3DSwinB) is included in this set selected using the RadioLung analysis.

Table \ref{tab:LUNAMetrics} reports Specificity, Sensitivity, F1Score, Accuracy and AUC in \textbf{HQ}, \textbf{LQ} scans for each of these selected methods on LUNA. We also include the overall metrics (labelled \textbf{ALL}) obtained in the whole population including missing metadata cases. Metrics and bootstrap 95\% CIs have been computed in oof predictions. 


For all predictors, Specificity, F1Score, Accuracy and AUC higher in \textbf{HQ}. Also the overall values \textbf{ALL} of these metrics are between  \textbf{HQ} and \textbf{LQ} ones. The largest difference among groups is in Specificity with average increase from 0.4 to 0.75 for LUNA worse performer according to the analysis of Table \ref{tab:RegressionModels_ByArch_Vertical_Final}. Sensitivity does not always obtain the highest values for the \textbf{HQ} group, with some methods achieving similar (or lower) performance in \textbf{HQ} compared to \textbf{LQ} condition. This can be attributed to the sample size of \textbf{HQ} malignant class (24 cases) rather than to the main effect associated to performance quality. Also \textbf{ALL} Sensitivity is not between \textbf{HQ} and \textbf{LQ} values, being the lowest one. Given the proportion of malign and benign cases in the 3 groups, we consider that this can not be attributed to a bias in predictions, but to a large number of low quality cases in 807 scans with missing metadata. 

\begin{table*}
\centering
\caption{Performance metrics for LUNA for each of quality (Q) groups. Malign/Benign samples for each group are: HQ(24/36), LQ(41/79) and ALL(343/644).}
\label{tab:LUNAMetrics}
\begin{tabular}{lllllll}
\toprule
\textbf{Method} & \textbf{Q}& \textbf{Specificity}  & \textbf{Sensitivity} & \textbf{Accuracy} & \textbf{F1Score} & \textbf{AUC} \\
\midrule
GLCM3D & HQ & [0.5  0.94] (0.72) & [0.55 1.0] (0.84) & [0.6  0.96] (0.78) & [0.6  0.97] (0.78) & [0.61 0.97] (0.79) \\
ViTB16 & LQ & [0.23 0.51] (0.37) & [0.73 1.0] (0.95) & [0.52 0.79] (0.65) & [0.49 0.74] (0.61) & [0.53 0.79] (0.66) \\
 & ALL & [0.64 0.75] (0.69) & [0.65 0.79] (0.72) & [0.66 0.75] (0.71) & [0.66 0.75] (0.71) & [0.66 0.75] (0.71) \\
  \midrule
GLCM3D& HQ & [0.5  0.94] (0.72) & [0.66 1.0] (0.93) & [0.64 1.0] (0.83) & [0.62 0.99] (0.81) & [0.65 1.  ] (0.83) \\
 Vgg16  & LQ & [0.33 0.64] (0.48) & [0.72 1.0] (0.93) & [0.56 0.83] (0.7) & [0.54 0.8 ] (0.67) & [0.56 0.83] (0.7) \\
 & ALL & [0.63 0.74] (0.69) & [0.69 0.84] (0.76) & [0.68 0.77] (0.73) & [0.68 0.77] (0.73) & [0.68 0.77] (0.73) \\
  \midrule
GLCM3D& HQ & [0.64 1.0] (0.82) & [0.63 1.0] (0.92) & [0.7 1.0] (0.88) & [0.69 1.0  ] (0.87) & [0.7 1.0] (0.87) \\
 SwinB  & LQ & [0.44 0.74] (0.59) & [0.63 1.0] (0.85) & [0.59 0.86] (0.72) & [0.58 0.84] (0.71) & [0.58 0.86] (0.72) \\
 & ALL & [0.71 0.81] (0.76) & [0.63 0.77] (0.7) & [0.69 0.78] (0.73) & [0.69 0.77] (0.73) & [0.69 0.78] (0.73) \\
  \midrule
GLCM3D & HQ & [0.53 0.96] (0.75) & [0.51 1.0] (0.8) & [0.6  0.96] (0.78) & [0.58 0.96] (0.77) & [0.6  0.95] (0.78) \\
MobileNetV3l & LQ & [0.25 0.56] (0.4) & [0.73 1.0] (0.94) & [0.54 0.8] (0.67) & [0.52 0.76] (0.64) & [0.54 0.81] (0.68) \\
 & ALL & [0.65 0.75] (0.7) & [0.68 0.83] (0.76) & [0.69 0.77] (0.73) & [0.69 0.77] (0.73) & [0.68 0.77] (0.73) \\
  \midrule
GLCM3D& HQ & [0.56 0.98] (0.77) & [0.55 1.0] (0.84) & [0.64 1.0] (0.82) & [0.62 0.98] (0.8) & [0.63 0.98] (0.81) \\
ConvNextB  & LQ & [0.35 0.67] (0.51) & [0.62 1.0] (0.84) & [0.54 0.8] (0.67) & [0.53 0.78] (0.66) & [0.53 0.8 ] (0.66) \\
 & ALL & [0.69 0.79] (0.74) & [0.63 0.77] (0.7) & [0.67 0.76] (0.72) & [0.67 0.76] (0.72) & [0.67 0.76] (0.72) \\
\bottomrule
\end{tabular}
\end{table*}


\section{Discussion}

\subsection{Parameter Adjustment}

Our stratification of parameter quality agrees with radiological and physical principles since early Pareto layers ($HQ_1$, $HQ_2$) correspond to higher X-ray current, thinner slice thickness and faster spiral pitch, while late ($HQ_{6}$, $HQ_{7}$) layers correspond to configurations with the lowest X-ray, thickest cuts and slowest pitch. An interesting observation is that slice thickness and spiral pitch can be relaxed and still produce {\bf HQ} scans provided that X-ray current increases,  as the configurations of the second $ \mathrm{HQ}_2$ and third $ \mathrm{HQ}_3$ Pareto shown in fig.\ref{fig:esqHQDef} layers illustrate.

The optimal configurations selected by our multi-objective statistical Pareto selection framework slightly differ from clinical guidelines. According to \cite{xu2021effect} the optimal protocol for small nodule diagnosis should acquire CT volumes using a helical scanning with spiral pitch set at 1:0 and a gantry rotation time of $<=0.5$. Also, devices should operate at a voltage (KVP) range of 100-140 kVp (recommended 120 kVp) with X-ray current (mA) from 100 to 400 mA and slice thickness $\leq$ 1 mm. Our study suggest that X-ray current and slice thickness are, in fact, the most influential parameters, though with different critical values. Table \ref{tab:RadioLungParetoFrontAnovaCompact} indicates that there is no extra benefit in increasing from 200 to 300 mA and also one configuration at 200 mA is included in the 1st Pareto layer. These results suggest that predictors might achieve good performance with smaller XRay current mA and suggests that low dose might be a feasible option for radiomic predictions. The differences in accuracy did not show a measurable reduction in predictive performance when slice thickness thresholds up to 1.25 mm were considered. Surprisingly, spiral pitch does not seem to play a critical role in performance as it can be increased to 1.5 without a reduction in performance.

LUNA performance degradation in Table \ref{tab:RegressionModels_ByArch_Vertical_Final} shows that all methods, except IntensityEfficientNetV2l, perform worse ($OR_n>1$) under the \textbf{LQ} conditions defined on the RadioLung dataset. However, IntensityEfficientNetV2l odds ratio is not significant and it is the worst performer of the LUNA set with a significant  $OR_{\textbf{HQ}_{n,0}}=4.2$. Given that a similar pattern is observed in other methods in both datasets (in particular, IntensityEfficientNetV2l and GLCM3DEfficientNetB7 in RadioLung), we conclude that for low accuracy predictors, the quality effect might not be significant. Since RadioLung was used to select optimal configurations based on similar GLMM models, it is expected to have higher odds ratios (as high as 84) in this set than in LUNA. This difference between over-optimistic train and more realistic test metrics can be also attributed to the different distribution of scan parameters (see fig.\ref{fig:DistParam}). Despite this decrease in odds ratios comparing \textbf{HQ} and \textbf{LQ} conditions, we consider that results obtained for the independent validation set LUNA support the transferability of the proposed statistical Pareto selection of parameters.

\subsection{Performance Sensitivity and Clinical Impact}

The comparison across radiomic methods reported in Section \ref{Sec:Exp2} suggests some interesting conclusions. First, methods using the original scan intensity domain might perform worse than models based on volumetric radiomic representations. Second, backbone architectures based on EfficientNet and ResNet models perform worse, especially on the intensity domain, with drops in performance rate up to 25 times worse in the case of IntensityEfficientNetB7. Analysis of top models indicates that the transformer ViTB16, hybrid models ConvNextB, SwinB and Vgg16 in radiomic domain and DenseNet169 in intensity domain are the backbones that achieve best performance for small and non-small nodules in \textbf{HQ} conditions. This suggests potentially stronger representation capacity of foundation models when combined with radiomic 3d representation domains. 

Results in \ref{Sec:ClinImpact} illustrate the clinical potential of our selection of acquisition parameters. There is over a 10\% increase in Accuracy, AUC and F1-scores, mainly coming from a substantial increase in specificity (over 20\%, regardless of the predictor). These findings support, the potential transportability of the performance sensitivity analysis. The lack of consistent increase in sensitivity suggests that maybe other co-variables (like malignancy prevalence) or effects (like reconstruction kernel, contrast protocol) should be considered to obtain stronger conclusions. 

\subsection{Limitations}

We are aware that OR estimates differences in the average accuracy of models, which might be considered a sub-optimal metric in case of unbalanced settings. Although other models could be formulated for assessing sensitivity to specific performance metrics, there are several reasons for this choice. Modeling acquisition sensitivity at the individual prediction level provides one outcome per case, allowing direct estimation of hierarchical generalized mixed-effects models with the full dataset sample size. In contrast, other performance metrics (like sensitivity, specificity, F1-score, or AUC) are aggregate quantities defined at subset level. Therefore, their use as dependent variables would require repeated data splitting or resampling to generate multiple observations, substantially reducing the effective sample size and making inference unstable, particularly for small datasets. Another reason for choosing OR is that models are naturally adjusted in logarithmic scale, thus, avoiding any hand-crafted transformation of the dependent variable required to satisfy normality and heteroscedasticity model assumptions. Finally, we would like to observe that in order to ensure that conclusions are clinically meaningful and not biased, Experiment 3 additionally evaluates sensitivity, specificity, F1-score, accuracy, and AUC under \textbf{HQ} and \textbf{LQ} conditions, showing consistent conclusions across metrics.

Regarding parameter space search, we used a grid search strategy to guarantee exhaustive exploration of the parametric space and given that computational burden of estimating GLMM is low. Once the predictive models have been trained the whole multi-objective search runs in few minutes, thus, allowing for low-cost exhaustive brute force search methods. Also, it is worth observing that, in real settings, the Pareto search does not explore a continuous parametric space, since it should only include feasible scan parameters values. This is scan-specific since it depends on the minimum increment in resolution allowed by the device. Therefore the binning resolution for the grid search discretization is bounded by the hardware specifications, which limits the number of parameter configurations and comparisons.

Selecting thresholds on a small cohort and then projecting them to a larger set might create uncertainty about whether the chosen region is not overfitted to the RadioLung dataset. Although LUNA is larger in size, the parameter distribution is narrower and does not properly cover the parametric space (as the boxplots in Figure \ref{fig:DistParam} show). This was the main reason for choosing RadioLung as a discovery set for the multi-objective parameter selection. We are aware that with 201 patients spread across
3 different centers, the small number of cases per site may introduce center-specific signatures, potentially leading to shortcut learning \cite{filiot2025distilling}. The visual analysis of the distribution of acquisition parameters in fig.\ref{fig:DistParam} does not indicate any bias or pattern that could be associated to either malignancy or institution (in the case of RadioLung). Although, this does guarantee absence of shortcut learning due to other confounding factors, we consider that, in view of the results for the validation LUNA set, this situation does not affect either the transferability of results or our main conclusions.


The size of RadioLung was not large enough for neither nested cross-validation, nor leave-one-center-out analysis. Given the available scans with full acquisition metadata, we used this set as discovery set for parameter space search and the scans with missing metadata as {\bf BLIND} validation set for the final selection of parameters. For the comparison across predictive models, RadioLung models were re-trained using a different 10-fold split. Even if this might mitigate training data variability, it does not provide an independent test set for performance sensitivity and model selection. This is why, RadioLung is only considered as a discovery cohort that is validated on the LUNA set. Results on this set indicate that the parameters selected on RadioLung also define \textbf{HQ} and \textbf{LQ} regions and that this split reflects an improvement in clinical metrics. 


\section{Conclusions}

This work presents a statistical multi-objective optimization framework for the assessment of radiomic systems to varying acquisition conditions and the selection of optimal parameter settings. By directly modeling acquisition-induced failure risk rather than relying on implicit correction through harmonization, this framework enables a principled and clinically interpretable assessment of model reliability under realistic imaging conditions. Rather than selecting a single optimal configuration, the method derives hierarchical sets of Pareto-optimal solutions and translates them into admissible regions defined by problem-specific inequality constraints. In this context, the proposed approach provides an interpretable, set-valued description of quality-preserving configurations, suitable for complex acquisition or system design problems. 

Although evaluated in CT lung cancer diagnosis, the framework itself is modality-agnostic and can be applied to any imaging-based AI system for which acquisition metadata are available. The proposed combination of mixed-effects modelling and multi-objective optimization provides a general strategy for identifying acquisition conditions associated with reliable and transferable AI performance across heterogeneous clinical environments.

The benefits of the proposed framework have been evaluated on several radiomic systems for malignancy diagnosis in CT scans. CT scans parameters have been adjusted on the own-collected RadioLung dataset and tested on the independent LUNA public dataset to assess across dataset reproducibility. Results on the LUNA set indicates that statistical inference should complement predictive models to improve a trustworthiness clinical usage. Validation results using two independent multicentre datasets provides evidence supporting the
transferability and potential benefits of the proposed parameter selection strategy. Results comparing radiomic methods also demonstrate that radiomic 3D representation domains exhibit reduced sensitivity to acquisition variability compared with intensity-based representations. Results also 
suggest a stronger representation capacity of foundation models based on transformers in comparison to convolutional classification models. Finally, comparison to clinical guidelines suggests that the impact of acquisition protocols in the performance of radiomic methods should be considered in clinical protocols in order to facilitate multicenter deployment.   

This work could be extended in several directions. First, we could explore the benefits of using alternative more efficient optimization methods, like differential evolution, hierarchical search or Bayesian approaches. Second, we would like to evaluate the generalizability of the framework in other imaging tasks, modalities, and clinical problems. Finally, we would also like to deploy the framework in clinical protocols to assess the benefits in clinical practice.
\bibliographystyle{cas-model2-names}

\bibliography{references_compact_elsevier_initials_etal_acronyms_journals}

@article{powell2009bobyqa,
  title={The BOBYQA algorithm for bound constrained optimization without derivatives},
  author={Powell, Michael JD and others},
  journal={Cambridge NA Report NA2009/06, University of Cambridge, Cambridge},
  volume={26},
  number={26-46},
  pages={1},
  year={2009}
}

@inproceedings{filiot2025distilling,
  title={Distilling foundation models for robust and efficient models in digital pathology},
  author={Filiot, Alexandre and Dop, Nicolas and Tchita, Oussama and Riou, Auriane and Dubois, R{\'e}my and Peeters, others},
  booktitle={International Conference on Medical Image Computing and Computer-Assisted Intervention},
  pages={162--172},
  year={2025},
  organization={Springer}
}

@article{MixedMdels,
  author = {Booth, J. G.},
  title = "{Generalized Linear Models with Random Effects: Unified Analysis via H-Likelihood }",
  journal = {Biometrics},
  year = {2007},
  volume = {63},
  number = {4},
  pages = {1296-1297}
 
}

@inproceedings{ligero2019selection,
  author = {Ligero, M. and Torres, G. and Sanchez, C. and others},
  title = {Selection of radiomics features based on their reproducibility},
  booktitle = {EMBC},
  year = {2019},
  pages = {403--408}
}

@article{xu2021effect,
  author = {Xu, Y. and Lu, L. and Sun, S. H. and others},
  title = {Effect of CT image acquisition parameters on diagnostic performance of radiomics in predicting malignancy of pulmonary nodules of different sizes},
  journal = {Eur. Radiol.},
  year = {2021},
  pages = {1--11}
}

@article{setio2017validation,
  author = {Setio, A. A. A. and Traverso, A. and De Bel, T. and others},
  title = {Validation, comparison, and combination of algorithms for automatic detection of pulmonary nodules in computed tomography images: the LUNA16 challenge},
  journal = {Med. Image Anal.},
  year = {2017},
  volume = {42},
  pages = {1--13}
}

@article{simonyan2014very,
  author = {Simonyan, K. and Zisserman, A.},
  title = {Very deep convolutional networks for large-scale image recognition},
  journal = {arXiv},
  year = {2014}
}

@article{sun2018radiomics,
  author = {Sun, Roger and Limkin, Elaine Johanna and Vakalopoulou, Maria and others},
  title = {{A radiomics approach to assess tumour-infiltrating CD8 cells and response
		to anti-PD-1 or anti-PD-L1 immunotherapy: an imaging biomarker, retrospective
		multicohort study}},
  journal = {Lancet Oncol.},
  year = {2018},
  volume = {19},
  number = {9},
  pages = {1180--1191}
}

@article{lofstedt2019gray,
  author = {L{\"o}fstedt, T. and Brynolfsson, P. and Asklund, T. and others},
  title = {Gray-level invariant Haralick texture features},
  journal = {PLOS ONE},
  year = {2019},
  volume = {14},
  number = {2},
  pages = {e0212110}
}

@inproceedings{sandler2018mobilenetv2,
  author = {Sandler, M. and Howard, A. and Zhu, M. and others},
  title = {Mobilenetv2: Inverted residuals and linear bottlenecks},
  booktitle = {CVPR},
  year = {2018},
  pages = {4510--4520}
}

@article{traverso2018repeatability,
  author = {Traverso, A. and Wee, L. and Dekker, A. and others},
  title = {Repeatability and reproducibility of radiomic features: a systematic review},
  journal = {Int. J. Radiat. Oncol. Biol. Phys.},
  year = {2018},
  volume = {102},
  number = {4},
  pages = {1143--1158}
}

@article{mackin2015measuring,
  author = {Mackin, D. and Fave, X. and Zhang, L. and others},
  title = {Measuring computed tomography scanner variability of radiomics features},
  journal = {Invest. Radiol.},
  year = {2015},
  volume = {50},
  number = {11},
  pages = {757--765}
}

@inproceedings{wei2023ctflow,
  author = {Wei, L. and Yadav, A. and Hsu, W.},
  title = {CTFlow: Mitigating Effects of Computed Tomography Acquisition and Reconstruction with Normalizing Flows},
  booktitle = {MICCAI},
  year = {2023},
  pages = {413--422}
}

@inproceedings{he2016deep,
  author = {He, K. and Zhang, X. and Ren, S. and others},
  title = {Deep residual learning for image recognition},
  booktitle = {CVPR},
  year = {2016},
  pages = {770--778}
}

@inproceedings{tan2019efficientnet,
  author = {Tan, M. and Le, Q.},
  title = {Efficientnet: Rethinking model scaling for convolutional neural networks},
  booktitle = {ICML},
  year = {2019},
  pages = {6105--6114}
}

@article{hosseini2023deep,
  author = {Hosseini, S. H. and Monsefi, R. and Shadroo, S.},
  title = {Deep learning applications for lung cancer diagnosis: a systematic review},
  journal = {Multimed. Tools Appl.},
  year = {2023},
  pages = {1--31}
}

@article{li2022impact,
  author = {Li, Y. and Reyhan, M. and Zhang, Y. and others},
  title = {The impact of phantom design and material-dependence on repeatability and reproducibility of CT-based radiomics features},
  journal = {Med. Phys.},
  year = {2022},
  volume = {49},
  number = {3},
  pages = {1648--1659}
}

@article{rinaldi2022reproducibility,
  author = {Rinaldi, L. and De Angelis, S. P. and Raimondi, S. and others},
  title = {Reproducibility of radiomic features in CT images of NSCLC patients: an integrative analysis on the impact of acquisition and reconstruction parameters},
  journal = {Eur. Radiol. Exp.},
  year = {2022},
  volume = {6},
  number = {1},
  pages = {2}
}

@article{foy2020harmonization,
  author = {Foy, J. J. and Al-Hallaq, H. A. and Grekoski, V. and others},
  title = {Harmonization of radiomic feature variability resulting from differences in CT image acquisition and reconstruction: assessment in a cadaveric liver},
  journal = {Phys. Med. Biol.},
  year = {2020},
  volume = {65},
  number = {20},
  pages = {205008}
}

@article{fortin2018harmonization,
  author = {Fortin, J.-P. and Cullen, N. and Sheline, Y. I. and others},
  title = {Harmonization of cortical thickness measurements across scanners and sites},
  journal = {NeuroImage},
  year = {2018},
  volume = {167},
  pages = {104--120}
}

@article{ibrahim2021application,
  author = {Ibrahim, A. and Refaee, T. and Leijenaar, R. T. and others},
  title = {The application of a workflow integrating the variable reproducibility and harmonizability of radiomic features on a phantom dataset},
  journal = {PLOS ONE},
  year = {2021},
  volume = {16},
  number = {5},
  pages = {e0251147}
}

@article{ibrahim2022impact,
  author = {Ibrahim, A. and Lu, L. and Yang, H. and others},
  title = {The impact of image acquisition parameters and ComBat harmonization on the predictive performance of radiomics: a renal cell carcinoma model},
  journal = {Appl. Sci.},
  year = {2022},
  volume = {12},
  number = {19},
  pages = {9824}
}

@article{refaee2022ct,
  author = {Refaee, T. and Salahuddin, Z. and Widaatalla, Y. and others},
  title = {CT reconstruction kernels and the effect of pre-and post-processing on the reproducibility of handcrafted radiomic features},
  journal = {J. Pers. Med.},
  year = {2022},
  volume = {12},
  number = {4},
  pages = {553}
}

@article{yadav2025comparative,
  author = {Yadav, A. and Welland, S. and Hoffman, J. M. and others},
  title = {A comparative analysis of image harmonization techniques in mitigating differences in CT acquisition and reconstruction},
  journal = {Phys. Med. Biol.},
  year = {2025},
  volume = {70},
  number = {5},
  pages = {055015}
}

@article{demirciouglu2025rethinking,
  author = {Demircio{\u{g}}lu, A.},
  title = {Rethinking feature reproducibility in radiomics: the elephant in the dark},
  journal = {Eur. Radiol. Exp.},
  year = {2025},
  volume = {9},
  number = {1},
  pages = {85}
}

@article{wennmann2025reproducible,
  author = {Wennmann, M. and Rotkopf, L. T. and Bauer, F. and others},
  title = {Reproducible Radiomics Features from Multi-MRI-Scanner Test--Retest-Study: Influence on Performance and Generalizability of Models},
  journal = {J. Magn. Reson. Imaging},
  year = {2025},
  volume = {61},
  number = {2},
  pages = {676--686}
}

@article{torres2022intelligent,
  author = {Torres, G. and Baeza, S. and Sanchez, C. and others},
  title = {An intelligent radiomic approach for lung cancer screening},
  journal = {Appl. Sci.},
  year = {2022},
  volume = {12},
  number = {3},
  pages = {1568}
}

@article{baeza2024radiomics,
  author = {Baeza, S. and Gil, D. and Sanchez, C. and others},
  title = {Radiomics and clinical data for the diagnosis of incidental pulmonary nodules and lung cancer screening: radiolung integrative predictive model},
  journal = {Arch. Bronconeumol.},
  year = {2024},
  volume = {60},
  pages = {S22--S30}
}

@data{data1972_2025,
  author = {Gil, D. and Rosell, A. and Sánchez Ramos, C. and others},
  title = {{RadioLung}},
  year = {2025},
  doi = {10.34810/data1972}
}

@inproceedings{nagpal2022comparative,
  author = {Nagpal, P. and Bhinge, S. A. and Shitole, A.},
  title = {A comparative analysis of ResNet architectures},
  booktitle = {SMART GENCON},
  year = {2022},
  pages = {1--8}
}

@inproceedings{todi2023convnext,
  author = {Todi, A. and Narula, N. and Sharma, M. and others},
  title = {Convnext: A contemporary architecture for convolutional neural networks for image classification},
  booktitle = {CISCT},
  year = {2023},
  pages = {1--6}
}

@article{zhang2022transformer,
  author = {Zhang, C. and Jiang, W. and Zhang, Y. and others},
  title = {Transformer and CNN hybrid deep neural network for semantic segmentation of very-high-resolution remote sensing imagery},
  journal = {IEEE Trans. Geosci. Remote Sens.},
  year = {2022},
  volume = {60},
  pages = {1--20}
}

@article{wu2020visual,
  author = {Wu, B. and Xu, C. and Dai, X. and others},
  title = {Visual transformers: Token-based image representation and processing for computer vision},
  journal = {arXiv},
  year = {2020}
}

@book{coello2007evolutionary,
  author = {Coello, C. A. C. and Lamont, G. B. and Veldhuizen, D. A. V.},
  title = {Evolutionary algorithms for solving multi-objective problems},
  publisher = {Springer},
  year = {2007}
}

\end{document}